\documentclass[preprint,12pt]{elsarticle}

\usepackage{amssymb}

\usepackage{graphicx}%
\usepackage{multirow}%
\usepackage{amsmath,amssymb,amsfonts}%
\usepackage{amsthm}%
\usepackage{mathrsfs}%
\usepackage[title]{appendix}%
\usepackage{xcolor}%
\usepackage{textcomp}%
\usepackage{manyfoot}%
\usepackage{booktabs}%
\usepackage{algorithm}%
\usepackage{algorithmicx}%
\usepackage{algpseudocode}%
\usepackage{listings}%

\usepackage{amsmath,amssymb}
\usepackage{graphicx}
\usepackage[labelsep=space]{caption}
\usepackage{subcaption}
\usepackage{hanging}  
\usepackage{comment}

\journal{Neurocomputing}

\begin{document}

\begin{frontmatter}



\title{Adversarial Detection by Approximation \\ of Ensemble Boundary}


\author[1]{Terry Windeatt}
\ead{t.windeatt@surrey.ac.uk}

\affiliation[1]{organization={University of Surrey},
           addressline={CVSSP}, 
            city={Guildford},
            state={Surrey},
            postcode={GU2 7XH}, 
            country={UK}}

\begin{abstract}
Despite being effective in many application areas, \textit{Deep Neural Networks (DNNs)} are vulnerable to being attacked. In object recognition, the attack takes the form of a small perturbation added to an image, that causes the \textit{DNN} to misclassify, but  to a human  appears no different.
Adversarial attacks lead to defences that are themselves subject to attack, and 
the attack/defence strategies provide important information about the properties of \textit{DNNs}. 
In this paper, a novel method of detecting adversarial attacks is proposed for an ensemble of \textit{Deep Neural Networks (DNNs)} solving two-class pattern recognition problems. The ensemble is combined using Walsh coefficients which are capable of approximating Boolean functions and thereby controlling the decision boundary complexity. The  hypothesis in this paper is that decision boundaries with high curvature allow adversarial perturbations to be found, but change the curvature of the decision boundary, which is then approximated in a different way by Walsh coefficients compared to the clean images. Besides controlling boundary complexity, the coefficients also measure the correlation with class labels, which 
 may aid in understanding the learning and transferability properties of \textit{DNNs}. While the experiments here use images, the proposed approach of modelling two-class ensemble decision  boundaries could in principle be applied to any application area.   
\end{abstract}


\end{frontmatter}


\section{Introduction}

Despite achieving excellent performance in many application areas, \textit{Deep Neural Networks (DNNs)} are known to be susceptible to being fooled \cite{Craighero}  \cite{Han}. This can take many forms in various applications but visually is most striking when an image is manipulated, so that the \textit{DNN} misclassifies although the required perturbation results in an image that appears no different to a human. The perturbation is computed in various ways and known as adversarial pattern generation or adversarial attack. Example images of this phenomenon can be found, for example in  \cite{He23}. Although the attack usually consists of a digital manipulation, it can also be physical \cite{Nguyen}.  Lack of susceptibility to adversarial attacks is known as adversarial robustness and is clearly a desirable property.

Besides having important practical implications, especially for security and safety-related applications \cite{He23} the problem of adversarial attacks has been extensively studied theoretically. However, the existence of adversarial examples appears still to be a mystery \cite{Han}.  Indeed, there is recent recognition that adversarial pattern generation can help explain how \textit{DNNs} learn, which would be important since there are many aspects of \textit{DNN} learning which are poorly understood and remain open research questions \cite{Ortiz}. The main idea is that an adversarial attack, which attempts to push a pattern on the other side of a decision boundary, gives a notion of how the boundary is behaving during learning. 

For those unfamiliar with the basics of Neural Networks a summary  can be found in \cite{Qamar} \cite{Mijwel}. This paper is concerned with adversarial attacks on \textit{DNNs} with application to object recognition. However, secure communication and image encryption has also been studied in the context of Reaction-Diffusion Neural Networks (\textit{RDNNs}), for example the problem of state estimation for \textit{RDNNs} subject to Denial of Service (\textit{DOS}) attacks \cite{Song23a}, and bipartite synchronization issue for
double-layer switched cooperation-competition Neural Networks \cite{Song23b}. Further, deception attacks have been investigated in the context of  adaptive finite-time resilient  control for non-linear   fractional order large scale systems (\textit{FOLSS}) \cite{Song23c}, and also \textit{KNN} graphs for extracting fault features in complex gearbox vibration signals \cite{Tao}. 

As a result of the practical issue of the threat to \textit{DNNs}, adversarial attacks and subsequent defences have become somewhat of a game. In fact it is possible to consider the adversarial attacks in the context of game theory \cite{Chivukula} \cite{Addesso20}. For an Advanced Persistent Threat (\textit{APT}) rivalry game, information is disclosed from a player’s strategy adjustments, and the defender and attacker are able to work out the best timing for strategy adjustments based on the solution of the game  \cite{Zhang23game}. In \cite{Addesso21} an adversarial formulation of the threat propagation problem relies on \textit{Kendall’s birth-death immigration model}, and in \cite{Lian}, the switched \textit{Stackelberg} game is applied to model the switched decision-making process between the attacker and controller.

If attack/defence strategies represent a game, at present it appears that the adversarial attackers are winning.  There is evidence that the current adversarial defences that claim to be robust can all be defeated, particularly for object recognition \cite{Tramer} \cite{He23}. The most successful defence is adversarial training, in which the original training set is enhanced with adversarial patterns. However, reducing the  computational cost of adversarial training to make it a practical solution is an open research question \cite{Han}.

Ensembles or multiple classifier systems are a well-recognised method of solving pattern recognition problems. The two main phases in ensemble design are classifier generation and combination \cite{Windeatt06}. Classifier generation (known as individual or base classifiers) has the aim of producing accurate yet diverse classifiers, although diversity  has remained an elusive concept from the early days of ensembles up to the present \cite{Wood} \cite{Windeatt06}. The combination phase is usually simple, such as a \textit{Majority Vote}  or \textit{Weighted Vote}. Weighted combination rules have been extensively investigated but there is no established strategy for computing the weights. Trainable rules have been developed, but it is not clear whether there is any advantage and the training parameters can be difficult to set. 

When applied to \textit{DNNs}, there is a belief that ensembles have limited usefulness, because single state-of-the-art \textit{DNNs} achieve remarkable performance by themselves without the need for combining diverse individuals \cite{He17}. 
However, some researchers still claim that ensemble methods can  improve adversarial detection \cite{Deng} \cite{Craighero}. 
A different approach to transferability,  compared to the approach proposed here, is taken in  \cite{Guo}, which   evaluates the \textit{transferability prediction difference} of two \textit{DNN} models and trains a  threshold using clustering methods. 
Another  approach in \cite{Szucs} is to combine detection with making the attacked model more robust, and to examine the  filtering capability of the detector.

In this paper, an ensemble of parallel two-class \textit{DNNs} is proposed for approximating the decision boundary of clean and adversarial images using Walsh spectral coefficients, and thereby detecting adversarial attacks.  Walsh coefficients were first proposed for pattern recognition in the 1970’s \cite{Tou}, although the assumption was that the classifier inputs were binary features rather than class labels as inputs to a combining rule.  Like \textit{Majority Vote}, only class labels are used for Walsh spectral coefficients. The only parameter to choose in the Walsh combination is the order of the  coefficients, which acts as a form of regularisation \cite{Tikhonov} and determines how well the ensemble boundary is approximated.

The main hypothesis in this paper is that individual \textit{DNN} two class classifiers in an ensemble respond differently to adversarial examples with respect to their success in predicting the correct class label. The result is that there are changes in the decision boundary, and using Walsh coefficients to approximate the boundary enables adversarial examples to be detected.  Main contributions are as follows:
\begin{enumerate}
    \item Proposes a novel method of adversarial detection, based on modelling the decision  boundary using Walsh coefficients for two class problems. 
    \item Supports the view that understanding the  phenomenon of adversarial detection is aided by the decision boundary curvature hypothesis. By measuring the correlation with class labels, Walsh coefficients also aid in  understanding the learning and transferability properties of \textit{DNNs}.
    \item Experimentally demonstrates how to gradually increase the complexity of decision  boundary approximation. Also proposes a Walsh coefficient measure  that  enables adversarial detection based on a simple threshold. Furthermore, experiments show that as complexity is increased there is a trade-off between detection and clean image recognition rate.
\end{enumerate}	

Note that the Conclusion in Section 5 includes limitations of the proposed method. In particular, superiority over other adversarial defence methods is not claimed, since thee is evidence that  defence methods can be defeated by an adaptive attack, if enough is known about defence details \cite{Tramer}. So a fair comparison needs to determine how difficult it is to defeat the proposed defence in comparison with other methods, which is beyond the scope of this study. 

Section 2 describes how Walsh coefficients may be used for the combining rule of a classifier ensemble and shows that \textit{Weighted Vote} is a special case of Walsh approximation. Section 3 discusses the theories of adversarial robustness, particularly those related to decision boundaries, and defines the adversarial attacks used in experiments in Section 4. Datasets and classifiers are described in Section 4.1, definition of terms in 4.2, experimental results in 4.3, discussion in 4.4 and link to code in 4.5. 

\section{Ensembles Combined using Walsh Coefficients}
\label{walsh}

In this paper,  \textit{DNNs} are used as base classifiers in an ensemble that solves two-class classification problems. It is assumed that there are \textit{N} parallel base classifiers, with final layer of each \textit{DNN} providing a binary classification. Let  \(X_{m}\)  be the \textit{N}-dimensional binary vector that represents the decisions of the \textit{N} \textit{DNN} classifiers for the  \(m^{th}\) pattern.  If the target label for the \(m^{th}\) pattern is denoted by 
\(\Omega_{m}\) then  \(\Omega_{m}=\varphi(X_{m})\)  where  \(m = 1 \ldots \mu\), \(\Omega_{m} \in\)\ \{0,1\} and \(\varphi\)  is the unknown Boolean function that maps \(X_{m}\) to \(\Omega_{m}\). Each element of \(X_{m}\) represents a vertex in the \textit{N}-dimensional binary hypercube.


\begin{table}
\small
  \caption{Rademacher-Walsh Polynomial Functions \\ (order \(N \le 4\))}
  \label{tab:walsh}
\centering

\begin{tabular}{l l }
\toprule
 \(j\) & \(\phi_{j}(X)\) \\ 
\midrule
\(1\) & \(1\)  \\
\(2\) & \(2x_{1}-1 \) \\
\(3\) & \(2x_{2}-1 \) \\
\(\vdots\) & \(\vdots\)  \\
\(N+1\) & \( 2x_{N}-1 \)  \\
\(N+2\) & \( (2x_{1}-1)(2x_{2}-1) \)  \\
\(N+3\) & \( (2x_{1}-1)(2x_{3}-1) \)  \\
\(\vdots\) & \(\vdots\)  \\
\(N+2+N-1\) & \( (2x_{1}-1)(2x_{N}-1) \)  \\
\(N+3+N-1\) & \( (2x_{2}-1)(2x_{3}-1) \)  \\
\(\vdots\) & \(\vdots\)  \\
\(N+2+N(N-1)/2\) & \( (2x_{N-1}-1)(2x_{N}-1) \)  \\
\(N+3+N(N-1)/2\) & \( (2x_{1}-1)(2x_{2}-1)(2x_{3}-1) \)  \\

\(\vdots\) & \(\vdots\)  \\
\(2^{N}\) & \( (2x_{1}-1)(2x_{2}-1) \hdots  (2x_{N}-1)\)  \\

\bottomrule
\end{tabular}

\end{table}

Following \cite{Tou}, the discrete probability density function can be approximated using \textit{Rademacher-Walsh (RW)} polynomials as orthogonal basis functions. Therefore, in the context of an ensemble the probability of occurrence of each of the \(\mu\) binary votes is computed. The \textit{RW} polynomials contain \(2^{N}\)  possible terms and are formed by taking products as shown in Table  \ref{tab:walsh} for \(N \le4 \)  (The full table of the \textit{RW} polynomials for \(N >4\), and an explanation of how they may be used, can be found  in \cite{Tou}). \textit{RW} discrete polynomial functions are orthogonal, satisfying the property that

\begin{equation}
\sum^{2^{N}}_{m=1}	\phi_{j}(X_{m}) = 
\begin{cases} 
2^{N} & \text{if } j = k\\
0 & \text{if } j \ne k
\end{cases}
\end{equation}

An approximation of the discrete probability density \(p\) using \(q\) basis functions and \(\mu\) patterns is given by
\begin{equation}
\hat{p} = \sum^{q}_{j=1} \chi_{j}\phi_{j}(X)
\end{equation}
where coefficients are given by 
\begin{equation}
\chi_{k} = \frac{1}{2^{N}\mu}  \sum^{\mu}_{i=1} \phi_{k}(X_{i})
\end{equation}

Spectral coefficients may also be interpreted as correlation with class labels \cite{Hurst}. The first order coefficients, \(j = 2 \ldots N+1\) in Table  \ref{tab:walsh} represent the correlation with the class label. Second and higher order coefficients represent correlation with the logic \textit{exclusive-OR} (\textit{xor} denoted \(\oplus\)) of the respective  coefficients. For example, if \(j = N+2\) in Table  \ref{tab:walsh}, the second order spectral coefficient \(s_{12}\) is given by

\begin{equation}
s_{12} = \chi_{5} = \frac{1}{2^{N}\mu}  \sum^{\mu}_{m=1} C(x_{m1} \oplus x_{m2}, \Omega_{m})
\end{equation}

\[\text{where } C(a,b) = 
\begin{cases} 
+1 & \text{if } a = b\\
-1 & \text{if } a \ne b
\end{cases}
\]

A simple example shows how to compute the three combinations of the second order contribution for a three-dimensional binary pattern \([x_{1},x_{2},x{_3}]=[0,1,1] \). With reference to Table  \ref{tab:walsh} and (2) and (3), for \([x_{1},x_{2}]\) the basis function is \((2x_{1}-1)(2x_{2}-1)=(-1)(1)=-1\). Similarly, for \([x_{1},x_{3}]\) the count is -1 but for \([x_{2},x_{3}]\) the count is +1. Adding the three contributions gives a count of -1 for that pattern. There are eight basis functions in total (from (1) with \(N=3)\), the remaining being \(0^{th}\), \(1^{st}\) and \(3^{rd}\) order. Pseudocode for an algorithm to compute the correlation for specified classifiers follows, and a link to the full code can be found in Section 4.5.
\\
\\
\hrule
\vspace{0.25cm}
\noindent
\textbf{Algorithm for Walsh coefficient computation of specified classifiers} 
\vspace{0.1cm}
\hrule
\vspace{0.25cm}
\noindent
{\small
\textbf{Computes the Walsh coefficient  component  \textit{W} 
 where \\ \textit{bftrs} is a matrix having \textit{np} columns of   binary\textit{(0 or 1) vectors} \\
 \textit{np} is the number of patterns;   \\
\textit{targtrs} is a binary target class vector of length \textit{np}; \\ \textit{inds} is a vector of length \textit{numcoeff} specifying rows of \textit{bftrs} \\ e.g \textit{inds}  = \textit{[1,3,4]} requests the Walsh component for exclusive-OR correlation between classifiers \textit{1,3,4}, which are the rows of \textit{bftrs} } \\
\(1 \) \hspace{1cm}   a = 0; b = 0; c = 0; d = 0; \textbf{\textit{*initialise }} \\ 
 2  \hspace{1cm}  \textbf{for} m = 1:np \textbf{\textit{ *do for all patterns}}\\ %
 3 \hspace{1cm} \hspace{1cm} sumx = 1;  \textbf{\textit{*initialise}}\\ %
 4  \hspace{1cm}   \hspace{1cm}  \textbf{for} j = 1:numcoeff  \textbf{\textit{*do for required order}}\\ 
 5    \hspace{1cm}   \hspace{1cm} \hspace{1cm}     sumx = sumx * (2*bftrs(inds(j),m)-1); \textbf{\textit{*correlation \(\pm1\) }}\\ %
 6  \hspace{1cm} \hspace{1cm}      \textbf{end}  \\
 - \hspace{1cm} \hspace{1cm} \textbf{\textit{*sum according to  class label and correlation }}  \\
 7   \hspace{1cm}  \hspace{1cm}   \textbf{if} targtr(m) == 1  \\ 
 8 \hspace{1cm}  \hspace{1cm} \hspace{1cm}       \textbf{if} sumx == 1  \textbf{then}  a = a + 1  \textbf{else}   c = c + 1 \\ 
10  \hspace{0.8cm} \hspace{1cm} \textbf{else}  \\
11   \hspace{0.8cm}   \hspace{1cm}  \hspace{1cm}    \textbf{if} sumx == 1 \textbf{then}  b = b + 1  \textbf{else}   d = d + 1 \\         
12  \hspace{0.8cm}  \hspace{1cm}    \textbf{end} \\
13  \hspace{0.8cm}  \textbf{end} \\
14 \hspace{0.8cm}  an = a/np; cn = c/np; bn= b/np; dn = d/np; \textbf{\textit{*divide by np}}\\ %
15  \hspace{0.8cm}  W = (an - cn) - (bn - dn); \textbf{\textit{*sum normalised correlations}} \\
}
\hrule 
\vspace{0.5cm}

To compute the probability estimate of two classes 
\(\omega_{1}\), \(\omega_{0}\) assume prior probability 
\(p(\omega_{1})\), \(p(\omega_{0})\)  is given by the number of patterns.  
A decision function \(d(X)\)  is formed by subtracting the probability estimate of the two classes, which is the probability density in (2) multiplied by prior probability:

\begin{equation}
d(X)=\hat{p}(X \mid \omega_{1})p(\omega_{1})-\hat{p}(X \mid \omega_{0})p(\omega_{0}) 
\end{equation}
After normalising (5), the Walsh decision probability is used in Section 4. Note that the two probability estimates in (5) contain the same divisor (\(2^{N}\)) in (3), so can be ignored when computing the decision function.

The decision function based on the Walsh coefficient combiner of order \textit{y} is defined as \(W_{y}\), so third order is \(W_{3}\), and contains the addition of contributions from \(W_{0}\), \(W_{1}\), \(W_{2}\). Note from Table \ref{tab:walsh}  that \(W_{0}\) basis function (j=1) is constant and does not depend on the pattern elements. So \(W_{0}\) is just the difference of the number of patterns in the two classes. \(W_{1}\) is similar to an ensemble weighted vote with weights set by correlation of individual classifiers with class label.

An ensemble combined by Walsh coefficients enables a complex decision boundary to be approximated by much simpler classifiers. This is demonstrated in  \cite{Windeatt19} for the artificial 2D circular dataset problem, in which simple individual classifiers under-fit, yet \(W_{3}\) provides a good approximation to the optimal Bayes boundary, while \textit{Majority Vote}  and \(W_{1}\) do not.

\section{Adversarial robustness}
\label{headings}

Adversarial examples were first demonstrated in  \cite{Szegedy} \cite{Biggio}. In this paper, the focus is limited to adversarial robustness, as applied to images. The intention of an adversarial attack is to perturb an image so that it is misclassified, yet to the human eye appears little or no different. It is an interesting topic theoretically, because it shows the limitations of machine learning applied to images. 
From a practical standpoint, for an application that is safety or security related a misclassification could incur a high cost e.g. a self-driving car, that misclassifies a pedestrian. It is not obvious why human invisible adversarial perturbations provide a successful attack. However, perhaps it is not surprising because \textit{DNNs} generally learn using backpropagation, which is not biologically plausible. Indeed, some have questioned whether a different learning algorithm would need to be developed, more in line with the way humans learn \cite{Lin} \cite{Han}.

In the context of images, there are many theories as to why \textit{DNNs} are not robust to adversarial pattern generation.    It is argued in \cite{Goodfellow} that gradient-based optimisation requires \textit{DNN} classifiers to be designed to be sufficiently linear, and that this local linearity causes susceptibility to adversarial examples and the transference to different classifiers. The theoretical analysis in  \cite{Shafahi} identifies fundamental bounds on the susceptibility of a classifier to adversarial attacks, and concludes that adversarial examples may be inevitable. 
Other theories relate robustness to features
\cite{Zhang22}. For example, in  \cite{Ilyas}, features in the final layer of the \textit{DNN} are shown to be split into robust and non-robust features, the former being aligned with the visual features used by humans. 
Although there are many theories of why \textit{DNNs} are not adversarially robust, the focus here is on the boundary theory.

\subsection{Boundary Theory of Adversarial Patterns}

Many theories, either explicitly or implicitly through the data distribution, link adversarial robustness to the decision boundary \cite{He23} \cite{Zhang23}. 
The analysis in  \cite{Schmidt} links adversarially robust generalisation to over-fitting, arguing that depending on the particular dataset distribution, more data may be needed. In  \cite{Fawzi}, by improving the robustness to random noise, a relationship between robustness and curvature of the decision boundary is established, suggesting that geometric constraints on the curvature of the decision boundary should be imposed. 
A visual illustration of adversarial patterns close to the decision boundary can be found in  \cite{Crecchi}.

The  hypothesis in this paper is that decision boundaries with high curvature allow adversarial perturbations to be found, but change the curvature of the decision boundary, which is then approximated in a different way by Walsh coefficients compared to the clean images. 
A well-established and useful model for understanding ensemble behaviour for two-class problems was formulated in  \cite{Tumer}, in which the assumption is made that the distribution of both classes is Gaussian. The relationship between the ensemble model  and Walsh coefficients was subsequently used for ensemble design \cite{Windeatt11} \cite{Windeatt13}. The conventional wisdom is that if a classifier is too complex, achieving a very low or zero training error rate, then the patterns under the tails of the distribution are correctly classified, leading to a decision boundary that differs from the optimal Bayes boundary, and generalises poorly through over-fitting. Apparently, a \textit{DNN} can produce a decision boundary capable of correctly classifying patterns under the tail, yet still perform well on the test set.    A possible reason is that high curvature of decision boundaries enables the patterns to be correctly classified, while other test error patterns in the vicinity are not misclassified. Imagine that each pattern in the overlap area is correctly classified via its own highly curved, possibly disconnected boundary. If the boundary is very tight, it would not affect generalisation, yet could be easily perturbed to misclassify. The visualisations of highly curved decision boundaries in  \cite{Ortiz}  \cite{Fawzi} suggest that this may be the case. The difficulty is that the space is high-dimensional and there is no guarantee that 2D visualisations are giving a true picture. It is better perhaps to think of it as a tool to try to help understand in an intuitive sense the reason for the relationship between accuracy and robustness.

\subsection{Adversarial Attacks/Defences}

Experiments described in Section 4 are based on checking adversarial detection for several adversarial attacks (\textit{ADVs})  that use a variety of attack methods. The \textit{ADVs} considered were \textit{Carlini-Wagner, ElasticNet, HopSkipJump, Zeroth Order Optimisation}, along with the five chosen below and  all described in   \cite{Nicolae}. In Section 4, \textit{dog/cat(d/c)} is chosen as the most difficult \textit{CIFAR10} \cite{Krizhevsky} problem in terms of error rate.  The following \textit{ADVs} used for experiments in Section 4 and implemented in  \cite{Nicolae}, were chosen based on their attack success on the \textit{d/c} dataset with default parameters selected: \textit{Deepfool (DF)}  \cite{Moosavi16}, \textit{Projection Gradient Descent (PG)} \cite{Madry}, \textit{Universal Adversarial Perturbation (UA)} \cite{Moosavi17}, \textit{Fast Gradient Sign Method (FG)} \textit{Basic Iterative Method (BI)} \cite{Kurakin}. While it was possible to increase attack success of the \textit{ADVs} by increasing perturbation levels, the attacks generally produced noticeable visual distortion of the images.

\textit{ADVs} may be classified according to attacker goal, knowledge and strategy \cite{Serban}. The goal is targeted versus untargeted, the former referring to an attack aimed at misclassifying a specified class, while the latter allows any misclassification.  Normally multi-class problems are assumed, but since only two class problems are considered in this paper, the attack is targeted, which is a more difficult problem \cite{Aldahdooh}.
Attacker knowledge is either white box or black box. White box assumes that the attacker has complete knowledge of the classifier and parameters, while black box relies on the transferability of attack images produced by a classifier that is in general different from the target classifier. All attacks considered in Section 4 are white box.  Attacker strategy is split into additive noise and geometric distortions (such as rotation and translation to induce misclassifications). All attacks considered here rely on additive noise, either using optimisation algorithms \textit{(PG, DF, UA)} or simpler and faster sensitivity analysis to determine the contribution of each input feature, for example \textit{gradient descent} or \textit{Jacobian matrix}. 

Adversarial defences are based on input transformation, adversarial detection or architectural design \cite{Serban}. A defence based on Walsh approximation would be considered an adversarial detection approach. It requires a rejection threshold to be set, similar to many defences, for example  \cite{Kwon},  which is based on finding a classification score threshold. 
Adversarial defences have in some cases been based upon ensembles \cite{Pang} \cite{Sen} \cite{Verma}. However, the consensus appears to be that ensemble defences, like other defences,  can be broken by adaptive attacks \cite{Serban} \cite{Tramer} \cite{He17}.

\section{Experimental evidence}

The experiments are designed to test the hypothesis that a Walsh coefficient approximation of an ensemble boundary allows a relatively weak \textit{DNN} classifier combination to detect  adversarial images. Although each base classifier is weak, the ensemble gives recognition rates higher than the average of individual classifiers and close to optimal for the datasets considered in this paper. In order to establish detection ability, the difference in approximation between clean and adversarial images is measured as Walsh coefficient order is increased. 

The aim is to find a measure based on Walsh coefficients that is able to distinguish between clean and adversarial images. The Walsh coefficient order \(W_{y}\) is defined in Section 2. As explained below, two measures are chosen \(W_{2}\) and the difference between \(W_{1}\) and \(W_{2}\) (\(W_{1}/W_{2}\)). A threshold for the Walsh decision probability (normalised from (5)) is found that optimises the accuracy of clean image acceptance and adversarial rejection. 


\subsection{Data-sets and classifiers}

Various two-class problems were chosen according to difficulty from \textit{CIFAR10} \cite{Krizhevsky}, \textit{dog/cat (d/c)}, \textit{truck/auto(t/a)}, \textit{horse/deer(h/d)}, \textit{frog/cat(f/c)}, \\
\textit{deer/bird(d/b)}  and from \textit{MNIST} digit recognition \cite{LeCun} \textit{9/4, 7/2, 9/7, 8/5, 8/3}. 

Most experiments use a twenty layer base classifier architecture \textit{NN20}. For comparison, two further architectures were chosen, a fourteen layer architecture \textit{NN14}, which is a subset of \textit{NN20} with layers 7-12 removed, and \textit{Alexnet} \cite{Krizhevsky12}.  Details can be found in Section 4.5.

Training uses gradient descent with \textit{adam} optimiser. The ensemble has twenty-one base classifiers, since it was found experimentally that increasing the number beyond twenty-one did not improve or decrease performance for the datasets and number of training patterns considered here. The diversity in each base classifier of the ensemble is due to random starting weights and drop-out. 

Classifier parameters were selected based on \textit{d/c}, which is the most difficult two-class problem in \textit{CIFAR10}, and then fixed for testing other two-class problems. The main parameters to set are the number of training epochs, number of nodes in the final layer and the drop-out rate. These were selected as 30, 512 and 50 percent respectively. It was found that varying the number of epochs/nodes and drop-out rate did not significantly alter performance. 

\subsection{Definition of terms}

The nomenclature for this section is as follows: Adversarial attacks \textit{ADVs} \textit{DF,PG,UA,FG,BI} are defined in Section 3 and use default parameters as given in  \cite{Nicolae}, except that for \textit{FG,BI} a value of twenty is used for epsilon.   \textit{TE} and \textit{TR}  refer to  the clean  test and training images, respectively, while \textit{ADVTE} and \textit{ADVTR} refer to the adversarial perturbed images e.g. \textit{DFTR} refers to \textit{DeepFool(DF)} training images. Class 1 and 2 are abbreviated as \textit{cl1} and \textit{cl2}, so for example \textit{TEcl1} refers to class 1 of  the clean test images. Note that  \textit{TEcl1} is compared with \textit{ADVcl2} since both \textit{TE} and \textit{ADV} images are predicted class 1. Similarly \textit{TEcl2} is compared with \textit{ADVcl1}. For experiments in this paper, the goal of adversarial detection is to maximise both the percentage of \textit{TE} images accepted, denoted \textit{TEACC} and the percentage of \textit{ADV}  images rejected, denoted \textit{ADVREJ}. 

\subsection{Experimental results for \textit{DNN} ensemble}

 Rather than use all images for the Walsh coefficient computation, in order to achieve the most reliable results, images that are incorrectly classified are removed.  
  Both \textit{TR} and \textit{TE} are split such that the clean image is retained if correctly classified by \textit{Majority Vote}. Similarly each \textit{ADV} is split so that the image is retained if incorrectly classified. Initially, for  training the classifiers the full training set is used, and then the Walsh coefficient computation is reported for the split images.
  
\begin{figure*}
     \centering
     \begin{subfigure}[b]{0.45\textwidth}
         \centering
     \includegraphics[width=\textwidth]{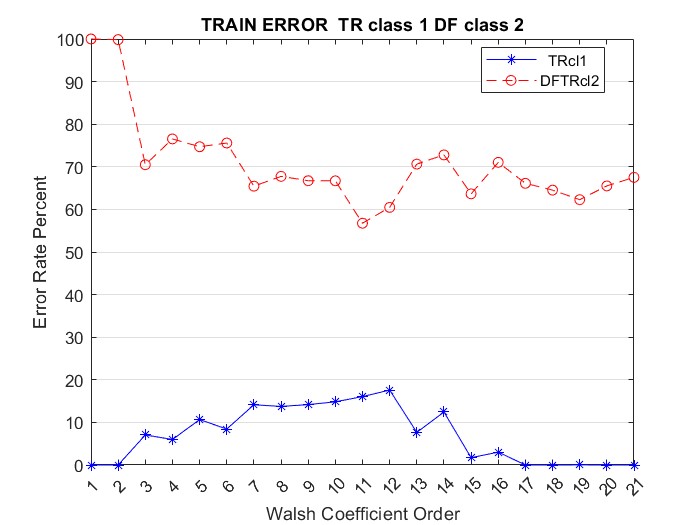}
         \caption{Class 1:\textit{dog} and Class 2:\textit{cat}}
         \label{fig: 1a}
     \end{subfigure}
     \hfill
     \begin{subfigure}[b]{0.45\textwidth}
         \centering
        \includegraphics[width=\textwidth]{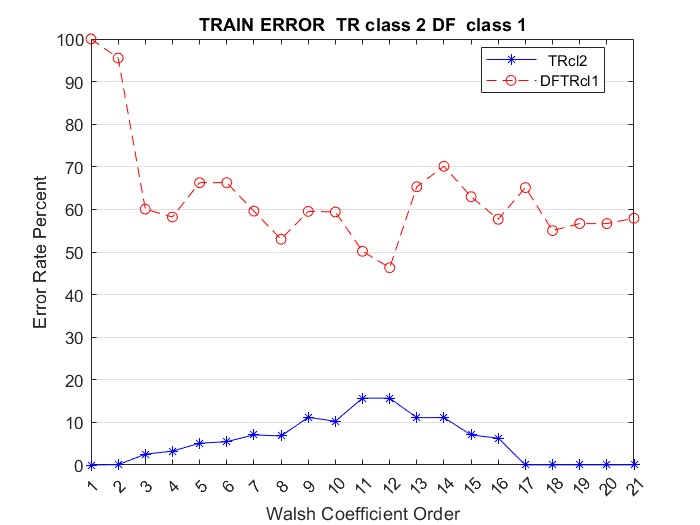}
         \caption{Class1:\textit{cat} and Class 2:\textit{dog}}
         \label{fig: 1b}
     \end{subfigure}
        \caption{Train error versus Walsh coefficient order for clean and \textit{Deepfool(DF)} training images }
        \label{fig: 1ab}
\end{figure*}

\begin{figure*}
     \centering
     \begin{subfigure}[b]{0.45\textwidth}
         \centering
     \includegraphics[width=\textwidth]{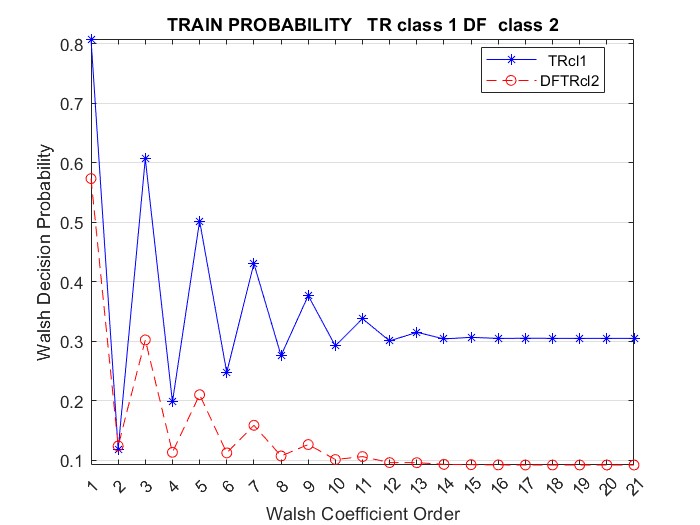}
         \caption{Class 1:\textit{dog} and Class 2:\textit{cat}}
         \label{fig: 2a}
     \end{subfigure}
     \hfill
     \begin{subfigure}[b]{0.45\textwidth}
         \centering
        \includegraphics[width=\textwidth]{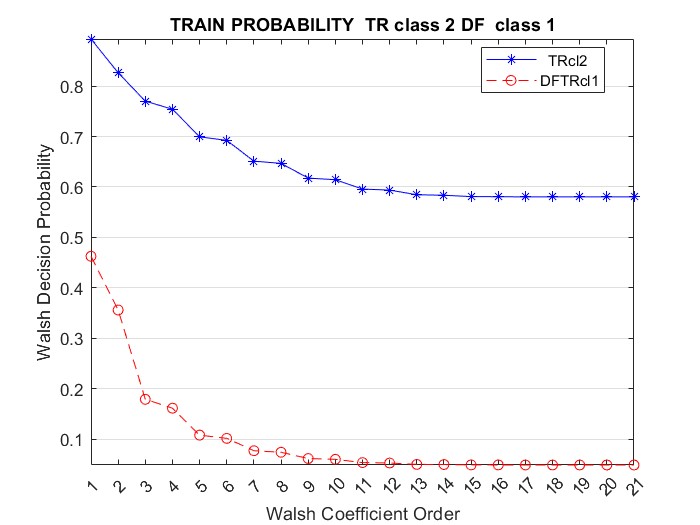}
         \caption{Class 1:\textit{cat} and Class 2:\textit{dog}}
         \label{fig: 2b}
     \end{subfigure}
        \caption{Walsh decision probability versus  coefficient order for clean and \textit{Deepfool(DF)} training images}
        \label{fig: 2ab}
\end{figure*}

\begin{figure*}
     \centering
         \begin{subfigure}[b]{0.45\textwidth}
         \centering
         \includegraphics[width=\textwidth]{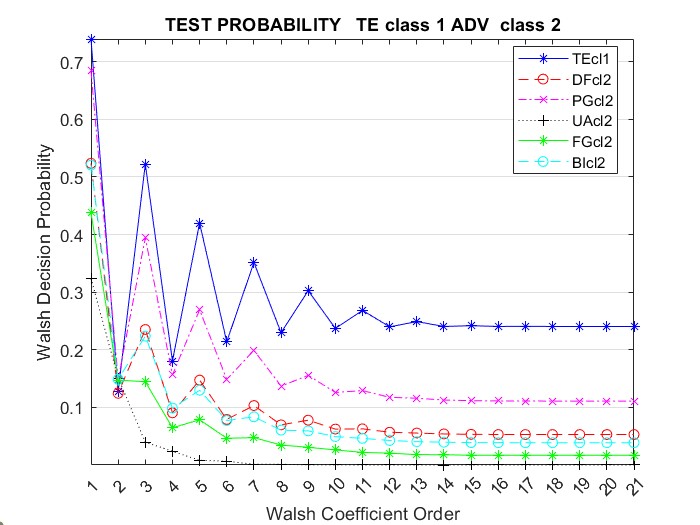}
         \caption{Class 1:\textit{dog} and Class 2:\textit{cat}}
         \label{fig: 2c}
     \end{subfigure}
     \hfill
        \begin{subfigure}[b]{0.45\textwidth}
         \centering
        \includegraphics[width=\textwidth]{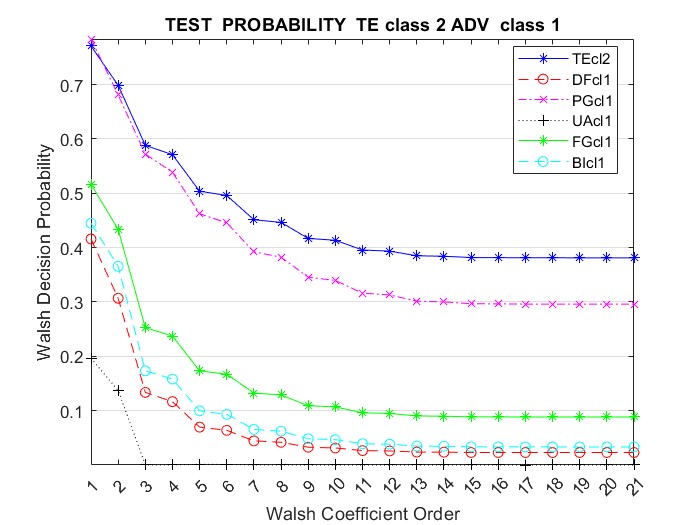}
         \caption{Class 1:\textit{cat} and Class 2:\textit{dog}}
         \label{fig: 2d}
     \end{subfigure}
        \caption{Walsh decision probability versus  coefficient order for Test and five \textit{ADVs} }
        \label{fig: 2cd}
\end{figure*}

\begin{figure}
         \centering
        \includegraphics[scale=0.3]{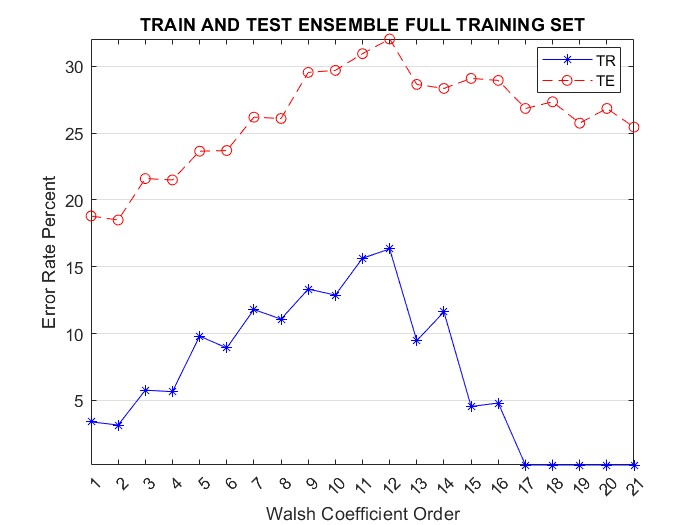}
         \caption{Ensemble train and test error of \textit{dog/cat} versus Walsh coefficient order for train/test set before splitting}
         \label{fig: 3b}
\end{figure}

\begin{figure}
     \centering
    \includegraphics[scale=0.3]{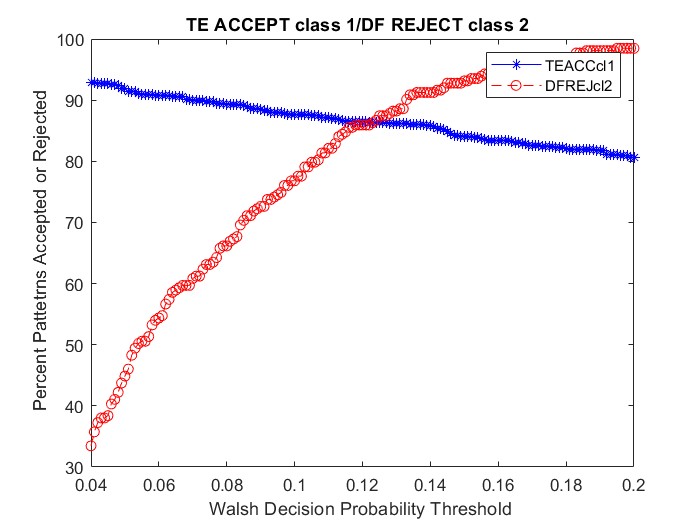}
         \caption{Typical TEACC/ADVREJ curve as \(W_{1}/W_{2}\)  detection threshold is varied}
         \label{fig: 3a}
\end{figure}

\begin{figure*}
     \centering
         \begin{subfigure}[b]{0.45\textwidth}
         \centering
         \includegraphics[width=\textwidth]{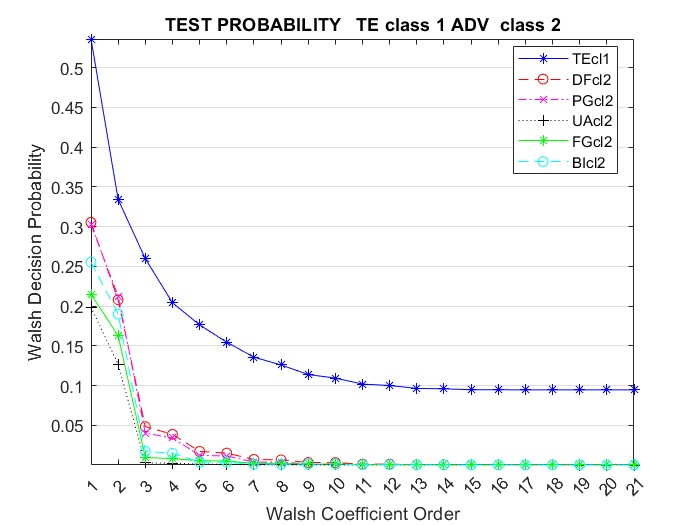}
         \caption{Class 1:\textit{dog} and Class 2:\textit{cat}}
         \label{fig: 6a}
     \end{subfigure}
     \hfill
        \begin{subfigure}[b]{0.45\textwidth}
         \centering
        \includegraphics[width=\textwidth]{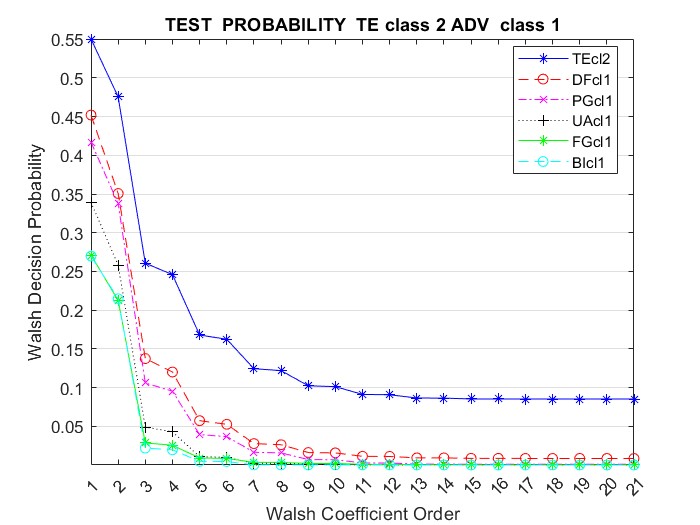}
         \caption{Class 1:\textit{cat} and Class 2:\textit{dog}}
         \label{fig: 6b}
     \end{subfigure}
        \caption{Walsh decision probability versus  coefficient order for Test and five \textit{ADVs} for \textit{NN14} base classifier}
        \label{fig: 6ab}
\end{figure*}

\begin{figure*}
     \centering
         \begin{subfigure}[b]{0.45\textwidth}
         \centering
         \includegraphics[width=\textwidth]{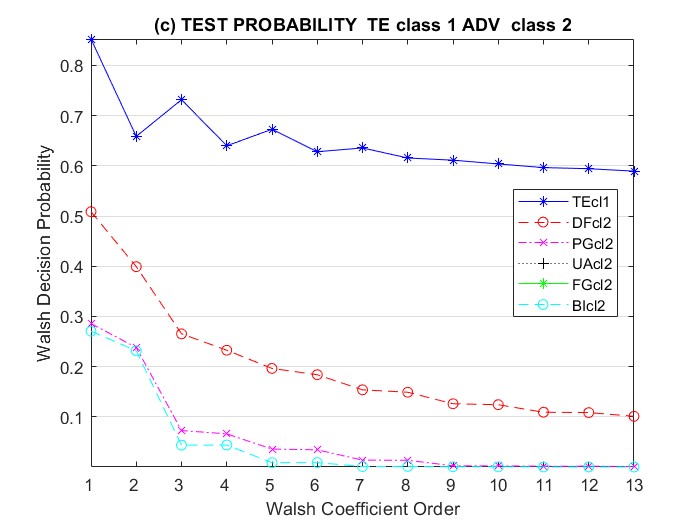}
         \caption{Class 1:\textit{dog} and Class 2:\textit{cat}}
         \label{fig: 7a}
     \end{subfigure}
     \hfill
        \begin{subfigure}[b]{0.45\textwidth}
         \centering
        \includegraphics[width=\textwidth]{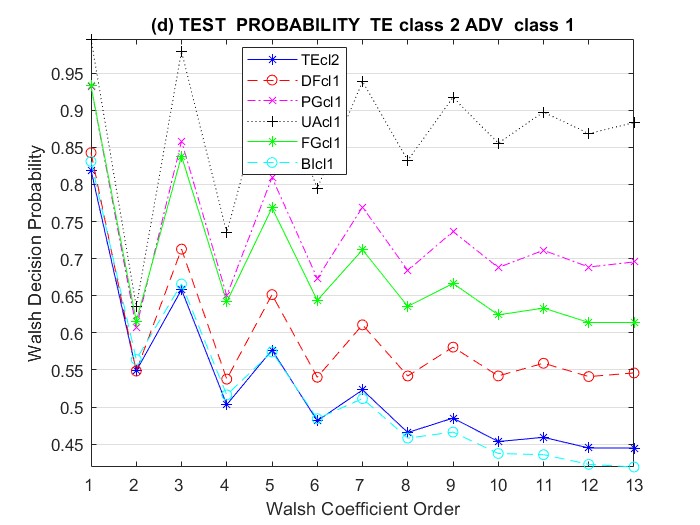}
         \caption{Class 1:\textit{cat} and Class 2:\textit{dog}}
         \label{fig: 7b}
     \end{subfigure}
        \caption{Walsh decision probability versus  coefficient order for Test and five \textit{ADVs} for Alexnet base classifier}
        \label{fig: 7ab}
\end{figure*}

Before a detailed analysis of results, a general summary follows.
The Figures in this section relate to \textit{dog/cat} problem, while Tables give results for all the \textit{CIFAR10} and \textit{MNIST} problems defined in Section 4.1. Figs. \ref{fig: 1ab}-\ref{fig: 3a}  and all Tables use \textit{NN20} architecture.   Figs. \ref{fig: 1ab}, \ref{fig: 2ab} show the difference between clean and \textit{Deepfool (DF)} training images, demonstrating the difference between training performance as Walsh order and therefore boundary complexity is increased.  The main result for adversarial detection is from Fig. \ref{fig: 2cd} which provides for \textit{NN20} the comparison of clean test images with  all \textit{ADVs} as Walsh order is increased.   Fig. \ref{fig: 3b} shows \textit{TR} and \textit{TE} ensemble error using the full training/test set, that is before they are split. Fig. \ref{fig: 3a} shows a typical acceptance/rejection curve  as the Walsh detection probability threshold is varied. Equivalent results for Fig. \ref{fig: 2cd} are provided for two different architectures,  Fig. \ref{fig: 6ab} for \textit{NN14} and Fig. \ref{fig: 7ab} for  \textit{Alexnet}, and demonstrate that even with different architectures it is possible to use Walsh detection probability to distinguish adversarial from clean images.

Figs. \ref{fig: 1ab}-\ref{fig: 2ab} show classification error and Walsh decision probability  as \(W_{y}\) is increased for \textit{d/c}, and demonstrate the difference between clean and adversarial training images (that is  between \textit{TR} and \textit{DFTR} and between \textit{TE} and \textit{DFTE}).
Fig. \ref{fig: 1a}  shows ensemble \textit{TRcl1} and \textit{DFTRcl2} average error versus order of Walsh coefficient. Fig. \ref{fig: 1b} is similar to Fig. \ref{fig: 1a}, showing error for \textit{TRcl2} and \textit{DFTRcl1}. Figs. \ref{fig: 2a}, \ref{fig: 2b} show the equivalent to Fig. \ref{fig: 1ab}  for Walsh decision probability. Note that for Fig. \ref{fig: 1ab}  the \(W_{1}\)\ \textit{TR} error is  0 while \textit{DFTR} error is 100 percent, since the data-set was split as explained above.

Figs. \ref{fig: 2c}, \ref{fig: 2d}  show  decision probability  for \textit{TE} versus \textit{ADVTE} for all five \textit{ADVs}.  For \textit{TEcl1} versus \textit{ADVcl2} Figs. \ref{fig: 2a}, \ref{fig: 2c} show that Walsh probability is generally lower for \textit{ADVcl2} with higher oscillations from odd to even coefficients. For \textit{TEcl2} vs \textit{ADVcl1} Figs. \ref{fig: 2b}, \ref{fig: 2d} show that Walsh probability is generally lower for \textit{ADVcl1} and oscillations not as noticeable. Curves of \textit{ADVs} in comparison with \textit{TE} shown in Figs. \ref{fig: 2c}, \ref{fig: 2d} show similar trends to Figs. \ref{fig: 2a}, \ref{fig: 2b} indicating that transferability across classifiers trained on \textit{ADVs} can be used to detect adversarial images. 

 From Figs. \ref{fig: 2c}, \ref{fig: 2d} there are various measures that could be used to potentially detect \textit{ADVs}. For the set of experiments presented in this paper the second order coefficient \(W_{2}\)  and the change from first to second order \(W_{1}/W_{2}\)   is chosen for each dataset and each class. Fig. \ref{fig: 3b} shows \textit{TR} and \textit{TE} ensemble error using the full training/test set, demonstrating that the lowest test error occurs for  \(W_{1}\) and \(W_{2}\). 
 
 The difference between \textit{TE} and \textit{ADV} in Figs. \ref{fig: 2c}, \ref{fig: 2d}  is an average, and in order to determine whether the difference can be used to detect \textit{ADVs}, Fig. \ref{fig: 3a} shows a typical \textit{TEACC/ADVREJ} curve as the \(W_{1}/W_{2}\)  detection threshold is varied. A pattern is accepted if its probability is greater than probability threshold. The determination of the optimal threshold (0.125), which gives  87\% for \textit{TEcl1/DFcl2}, has not been investigated here. If \textit{TR} was used to find the optimal value in this example, the  threshold would be 0.15 giving \textit{TEcl1/DFcl2}  85/92\%. In practice, by using a validation set to find the optimal threshold, it should be possible to find a value between 87/87 and 85/92. By varying the threshold it would also be possible to alter the trade-off, for example from Fig. \ref{fig: 3a} \textit{DFREJ} = 100\% could potentially be achieved at \textit{TEACC} = 80\%. 
 
Tables \ref{table2}-\ref{table5} show \textit{TEACC} and \textit{ADVREJ} for all problems considered from \textit{CIFAR10} and \textit{MNIST}. Numbers in the tables are italicised to indicate that  \(W_{2}\) is chosen rather than \(W_{1}/W_{2}\).  No entry in the tables occurs when there are too few patterns that have been successfully attacked for a reliable estimate. 

Table \ref{table2} shows the optimal \textit{TEACC/ADVREJ} for all datasets in \textit{CIFAR10} using \textit{TEcl1/ADVcl2} and \textit{TEcl2/ADVcl1}. Table \ref{table3}  shows the values of \textit{TEACC/ADVREJ}  if the threshold is set by \textit{DF} (for all datasets except \textit{f/c}). From Table \ref{table3} it may be observed that the optimal thresholds for each \textit{ADV} are similar,  indicating that transferability of attack across classifiers is not strongly dependent on the particular attack. Table \ref{table4}  shows the optimal values if \textit{Majority Vote} is used for \textit{TEcl1/ADVcl2}. In comparison with Table \ref{table2}, the averages are much lower and there is much more variability among the different \textit{ADVs}. 

Table \ref{table5}  shows the equivalent results for \textit{MNIST} for \textit{TEcl1/ADVcl2} and \textit{TEcl2/ADVcl1}.  However this is an easier data-set allowing perfect separation so that the optimal \textit{TEACC/ADVREJ} for both class 1 and 2 may be obtained for all problems. The entries show the lower probability threshold and range over which both \textit{TEACC} and \textit{ADVREJ} are greater than 98 percent. It may be seen that the initial probability thresholds are very similar with high probability range (from 0.12 to 0.38 for  \textit{TEcl1/ADVcl2} and .15 to .52 for \textit{TEcl2/ADVcl1}), indicating that it should be possible to locate the optimal \textit{TEACC/ADVREJ}.

\subsection{Discussion of Results }
The main result in Fig. \ref{fig: 2cd} demonstrates how the clean and adversarial test images behave differently as Walsh order, and therefore decision boundary complexity, is increased. From Fig. \ref{fig: 2cd} the novel method for detecting adversarial images is derived, that is using 
  \(W_{2}\) and the difference between \(W_{1}\) and \(W_{2}\) (\(W_{1}/W_{2}\)). As shown in Section 2, \(W_{2}\) may be interpreted as correlation (exclusive-OR in (4))  between pairs of class labels. 

From Fig.  \ref{fig: 3b} it may be seen that as the Walsh coefficient order is increased from \(W_{12}-W_{21}\)  the generalisation becomes worse,  so this is classic over-fitting as the decision boundary becomes more complex and \textit{TR} can be learned perfectly.   The best generalisation is found for \(W_{1}\), \(W_{2}\)   which is the reason why these were chosen for the experimental results shown in Tables \ref{table2} to \ref{table5}. 
However from Fig. \ref{fig: 2cd}  there is much better separation of the adversarial  attacks at the higher order; so if higher order coefficients were chosen for the experiments,  better separation on the adversarial attacks would be expected but at the cost of reduced \textit{TE} classification accuracy. In Fig. \ref{fig: 3b}, for \(W_{21}\)  there is a loss of TE accuracy of approx. 6 percent compared with \(W_{1}\). From the results in this section, it appears that the choice of Walsh coefficient order   for optimal generalisation, may in general be different to that chosen for optimal adversarial robustness.  


Results shown in Fig. \ref{fig: 6ab} for \textit{NN14} and in Fig. \ref{fig: 7ab} for  \textit{Alexnet},  demonstrate that even with different architectures the trend is similar to Fig. \ref{fig: 2cd}, and it is possible to use Walsh detection probability to distinguish adversarial from clean images. Further results for \textit{NN14} and \textit{Alexnet} are not shown here, but demonstrate similar trends to Figs. 
\ref{fig: 1ab}, \ref{fig: 2ab} and \ref{fig: 3b}.

\subsection{Code Availability and classifier architecture}
Code for approximating Boolean functions using Walsh coefficients that includes computation of the decision function in (5), along with 20 layer classifier architecture, \textit{NN20}  used in the experiments available:  \\https://doi.org/10.24433/CO.3695905.v1.

\section{Conclusion}

The main advantage of using Walsh approximation of ensemble classifier decisions is that the complexity of the decision boundary can be controlled.
This is important for adversarial detection since, as explained in Section 3, one  goal of adversarial perturbation applied to images is to push a pattern on the other side of a decision boundary, but not too far from the boundary, which  could result in human-observable differences in the image.   

Only three relatively weak base classifiers have been studied in this paper, and  for the  use of  stronger \textit{DNN} base  classifiers,  there would be a need to increase diversity for the ensemble to be effective. Understanding the trade-off between accuracy, diversity and adversarial detection when using Walsh coefficients as an ensemble combining rule has not been considered in this paper.

In practice, as explained in Section 1,   adversarial defences can be shown to fail assuming enough is known about defence details. In order to claim superiority over other  defence methods, it would be necessary to compare difficulties of defeating the defences, and is beyond the scope of this paper. Further work is required to determine how difficult it is to defeat a defence based on approximating an ensemble decision boundary using Walsh coefficients. 
There has been no attempt to defeat it using an adaptive attack, that is one which is specifically designed to overcome the Walsh defence. A suggested way of overcoming such a defence is to design a loss function based on all classifiers in the ensemble, as discussed in  \cite{Tramer}, for the ensemble defences \cite{Sen} \cite{Verma}.  

Walsh coefficients may be interpreted as correlation between class labels as well as controlling decision boundary complexity,  and therefore aid in understanding the training and transferability properties of \textit{DNN} classifiers. Also it has been shown experimentally that there is similar behaviour across different adversarial attacks, and for two-class problems the two classes may show different properties. Further investigation is needed to understand theoretically the relationship between correlation and decision boundary complexity as measured by the Walsh approximation.

A further limitation of this study is  that it is restricted to images. However, while the experiments here use images, the proposed approach of modelling two-class ensemble decision  boundaries could in principle be applied to any application area. Further work will be aimed at carrying out experiments on more diverse datasets including pilot studies on wider applications. 

The analysis in this paper is restricted to two-class problems, but it may be easier to understand \textit{DNNs} using two-class, before considering the more complex multi-class case. Further work will be aimed at scaling the approach to multi-class using \textit{Error-Correcting-Output-Codes (ECOC)} \cite{Verma}, which reduces multi-class problems to two class. 

The topic of adversarial robustness highlights an important difference between machine and human learning. Further study is warranted into using Walsh coefficients for combining classifier decisions to model the ensemble decision boundary for a better understanding of \textit{DNN} learning. 
\\
\\
\\
\textbf{DECLARATIONS}  
\\
\\
\textbf{Conflict of interest} The author declares that there is  no conflict of competing interests.


\begin{table}
  \caption{CIFAR10 optimal  \textit{TEACC/ADVREJ} each entry: \\
 \textit{TECL1/ADVCL2 \(|\) TECL2/ADVCL1}}
  \label{table2}
\centering
\begin{tabular}{l l l l l l}
\toprule
\textit{Pro} & \textit{DF} & \textit{PG} & \textit{UA} & \textit{FG} & \textit{BI} \\
\midrule
d/c & 87\(|\)\textit{93} & 85\(|\)\textit{94} & 90\(|\)\textit{87} & 85\(|\)\textit{86} & 85\(|\)\textit{85} \\
t/a & \textit{93}\(|\)93 & \textit{89}\(|\)94 & ---\(|\)95 & \textit{91}\(|\)94 & \textit{85}\(|\)94 \\
h/d & 94\(|\)92 & 96\(|\)91 & 96\(|\)- & 96\(|\)79 & 95\(|\)78 \\
f/c & ---\(|\)94 & ---\(|\)92 & ---\(|\)95 & \textit{95}\(|\)93 & \textit{93}\(|\)93 \\
d/b & \textit{91}\(|\)94 & \textit{91}\(|\)95 & \textit{92}\(|\)-- & \textit{93}\(|\)92 & \textit{92}\(|\)94 \\
\midrule
Ave & 91\(|\)93 & 90\(|\)93 & 93\(|\)92 & 92\(|\)89 & 90\(|\)89 \\
\bottomrule

\end{tabular}

\end{table}

\begin{table}
  \caption{CIFAR10 optimal \textit{TEACC/ADVREJ} using \textit{DF} probability threshold (except f/c)  \\
 top 5 rows:\textit{ TECL1/ADVCL2} \\ bottom 5 rows: \textit{TECL2/ADVCL1}
}
  \label{table3}
\centering

\begin{tabular}{l l l l l l}
\toprule
\textit{Pro} & \textit{DF} & \textit{PG} & \textit{UA} & \textit{FG} & \textit{BI} \\
\midrule
d/c & 87 & 85/85 & 86/99 & 85/85 & 86/84 \\
t/a & 93 & 93/71 & - & 93/78 & 91/55 \\
h/d & 94 & 94/98 & 94/98 & 94/98 & 94/96 \\
f/c & - & - & - & 95 & 95/90 \\
d/b & 91 & 91/91 & 90/100 & 91/98 & 91/97 \\
\midrule
d/c & 93 & 93/95 & 86/98 & 87/85 & 86/84 \\
t/a & 93 & 93/99 & 94/97 & 93/96 & 93/95 \\
h/d & 92 & 91/91 & - & 78/88 & 75/90 \\
f/c & 94 & 94/90 & 94/99 & 94/92 & 94/92 \\
d/b & 94 & 94/99 & - & 94/91 & 94/94 \\
\bottomrule

\end{tabular}

\end{table}

\begin{table}
  \caption{CIFAR10 optimal \textit{TECL1/ADVCL2} using \textit{majority vote}}
  \label{table4}
\centering

\begin{tabular}{l l l l l l}
\toprule
\textit{Pro} & \textit{DF} & \textit{PG} & \textit{UA} & \textit{FG} & \textit{BI} \\
\midrule
d/c & 74 & 55 & 85 & 65 & 70 \\
t/a & 90 & 85 & - & 85 & 80 \\
h/d & 82 & 60 & 55 & 75 & 72 \\
f/c & - & - & - & 85 & 80 \\
d/b & 80 & 70 & 77 & 84 & 80 \\
\midrule
Ave & 81 & 67 & 72 & 79 & 76 \\
\bottomrule

\end{tabular}

\end{table}

\begin{table}
  \caption{ MNIST probability/RANGE x 100  
for > \textit{TEACC/ADVREJ} OF 98/98 PERCENT 
\\ top 5 rows: \textit{TECL1/ADVCL2} \\ bottom 5 rows: \textit{TECL2/ADVCL1}
}
  \label{table5}
\centering

\begin{tabular}{l l l l l l}
\toprule
\textit{Pro} & \textit{DF} & \textit{PG} & \textit{UA} & \textit{FG} & \textit{BI} \\
\midrule
9/4 & 2/12 & 2/12 & 3/12 & 2/13 & 2/13 \\
7/2 & 5/24 & 5/25 & 5/24 & 5/24 & 5/25 \\
9/7 & \textit{62}/\textit{38} & \textit{62}/\textit{38} & \textit{62}/\textit{38} & \textit{62}/\textit{38} & \textit{62}/\textit{38} \\
8/5 & 8/37 & 8/37 & - & 6/38 & 8/37 \\
8/3 & \textit{7/28} & \textit{7/28} & \textit{7/34} & \textit{7/28} & \textit{7/28} \\
\midrule
9/4 & \textit{9/15} & \textit{9/15} & \textit{8/22} & \textit{9/15} & \textit{9/15} \\
7/2 & \textit{6/39} & \textit{7/35} & \textit{4/61} & \textit{6/39} & \textit{6/44} \\
9/7 & 6/32 & 6/32 & - & 5/32 & 5/34 \\
8/5 & \textit{48/52} & \textit{48/52} & \textit{48/52} & \textit{48/52} & \textit{48/52} \\
8/3 & 4/20 & 3/21 & 4/20 & 4/20 & 4/21 \\
\bottomrule

\end{tabular}

\end{table}






\begin{thebibliography}{41}

\bibitem{Craighero}
Craighero, F., Angaroni, F., Stella, F., Damiani, C., Antoniotti, M., \& Graudenzi, A. (2023). Unity is strength: Improving the detection of adversarial examples with ensemble approaches. Neurocomputing, 554, 126576.
\bibitem{Han}
Han, S., Lin, C., Shen, C., Wang, Q., \& Guan, X. (2023). Interpreting adversarial examples in deep learning: A review. ACM Computing Surveys, 55(14s), 1-38.
\bibitem{He23}
He, K., Kim, D. D., \& Asghar, M. R. (2023). Adversarial machine learning for network intrusion detection systems: A comprehensive survey. IEEE Communications Surveys \& Tutorials, 25(1), 538-566.
\bibitem{Nguyen}
Nguyen, K., Fernando, T., Fookes, C., \& Sridharan, S. (2023). Physical Adversarial Attacks for Surveillance: A Survey. IEEE Transactions on Neural Networks and Learning Systems.
\bibitem{Ortiz}
Ortiz-Jiménez, G., Modas, A., Moosavi-Dezfooli, S. M., \& Frossard, P. (2021). Optimism in the face of adversity: Understanding and improving deep learning through adversarial robustness. Proceedings of the IEEE, 109(5), 635-659.
\bibitem{Qamar}
Qamar, R., \& Zardari, B. A. (2023). Artificial neural networks: An overview. Mesopotamian Journal of Computer Science, 2023, 124-133.
\bibitem{Mijwel}
Mijwel, M. M., Esen, A., \& Shamil, A. (2023). Overview of neural networks. Babylonian Journal of Machine Learning, 2023, 42-45.
\bibitem{Song23a}
Song, X., Wu, N., Song, S., \& Stojanovic, V. (2023). Switching-like event-triggered state estimation for reaction–diffusion neural networks against DoS attacks. Neural Processing Letters, 55(7), 8997-9018.
\bibitem{Song23b}
Song, X., Wu, N., Song, S., Zhang, Y., \& Stojanovic, V. (2023). Bipartite synchronization for cooperative-competitive neural networks with reaction-diffusion terms via dual event-triggered mechanism. Neurocomputing, 550, 126498.
\bibitem{Song23c}
Song, X., Sun, P., Song, S., \& Stojanovic, V. (2023). Finite-time adaptive neural resilient DSC for fractional-order nonlinear large-scale systems against sensor-actuator faults. Nonlinear dynamics, 111(13), 12181-12196.
\bibitem{Tao}
Tao, H., Shi, H., Qiu, J., Jin, G., \& Stojanovic, V. (2023). Planetary gearbox fault diagnosis based on FDKNN-DGAT with few labeled data. Measurement Science and Technology, 35(2), 025036.
\bibitem{Chivukula}
Chivukula, A. S., Yang, X., Liu, W., Zhu, T., \& Zhou, W. (2020). Game theoretical adversarial deep learning with variational adversaries. IEEE Transactions on Knowledge and Data Engineering, 33(11), 3568-3581.
\bibitem{Addesso20}
Addesso, P., Cirillo, M., Di Mauro, M., \& Matta, V. (2020). ADVoIP: Adversarial Detection of Encrypted and Concealed VoIP. IEEE Transactions on Information Forensics and Security, 15, 943-958.
\bibitem{Zhang23game}
Zhang, L., Zhu, T., Hussain, F.K., Ye, D., \& Zhou, W. (2023). A Game-Theoretic Method for Defending Against Advanced Persistent Threats in Cyber Systems. IEEE Transactions on Information Forensics and Security, 18, 1349-1364.
\bibitem{Addesso21}
Addesso, P., Barni, M., Di Mauro, M., \& Matta, V. (2021). Adversarial Kendall’s Model Towards Containment of Distributed Cyber-Threats. IEEE Transactions on Information Forensics and Security, 16, 3604-3619.
\bibitem{Lian}
Lian, J., Jia, P., Wu, F., \& Huang, X. (2023). A Stackelberg Game Approach to the Stability of Networked Switched Systems Under DoS Attacks. IEEE Transactions on Network Science and Engineering, 10, 2086-2097.
\bibitem{Tramer}
Tramer, F., Carlini, N., Brendel, W., \& Madry, A. (2020). On adaptive attacks to adversarial example defenses. Advances in neural information processing systems, 33, 1633-1645.
\bibitem{Windeatt08}
Windeatt, T. (2008). Ensemble MLP classifier design. In Computational Intelligence Paradigms: Innovative Applications (pp. 133-147). Berlin, Heidelberg: Springer Berlin Heidelberg.
\bibitem{Wood}
Wood, D., Mu, T., Webb, A. M., Reeve, H. W., Lujan, M., \& Brown, G. (2023). A unified theory of diversity in ensemble learning. Journal of Machine Learning Research, 24(359), 1-49.
\bibitem{Windeatt06}
Windeatt, T. (2006). Accuracy/diversity and ensemble MLP classifier design. IEEE Transactions on Neural Networks, 17(5), 1194-1211.
\bibitem{He17}
He, W., Wei, J., Chen, X., Carlini, N., \& Song, D. (2017). Adversarial example defense: Ensembles of weak defenses are not strong. In 11th USENIX workshop on offensive technologies (WOOT 17).
\bibitem{Deng}
Deng, Y., \& Mu, T. (2024). Understanding and improving ensemble adversarial defense. Advances in Neural Information Processing Systems, 36.
\bibitem{Guo}
Guo, F., Zhao, Q., Li, X., Kuang, X., Zhang, J., Han, Y., \& Tan, Y. A. (2019). Detecting adversarial examples via prediction difference for deep neural networks. Information Sciences, 501, 182-192.
\bibitem{Szucs}
Szűcs, G., \& Kiss, R. (2023). 2N labeling defense method against adversarial attacks by filtering and extended class label set. Multimedia Tools and Applications, 82(11), 16717-16740.
\bibitem{Tou}
Tou, J.T., Gonzales, R.C. (1974) \textit{Pattern Recognition Principles}, Addison-Wesley, pp 151-4.
\bibitem{Tikhonov}
Tikhonov AN, Arsenin VA (1977) \textit{Solutions of Ill-posed Problems},   Winston \& Sons, Washington, USA.
\bibitem{Hurst}
Hurst L, Miller DM,  Muzio J (1985) \textit{Spectral Techniques in Digital Logic}, Academic Press, NY, USA,.
\bibitem{Windeatt19}
Windeatt, T., Zor, C., \& Camgoz, N. C. (2018). Approximation of Ensemble Boundary Using Spectral Coefficients. IEEE Transactions on Neural Networks and Learning Systems, 30(4), 1272-1277.
\bibitem{Szegedy}
Szegedy C., (2014) \textit{Intriguing properties of neural networks},  arXiv:1312.6199v4.
\bibitem{Biggio}
Biggio, B., Corona, I., Maiorca, D., Nelson, B., Šrndić, N., Laskov, P., ... \& Roli, F. (2013). Evasion attacks against machine learning at test time. In Machine Learning and Knowledge Discovery in Databases: European Conference, ECML PKDD 2013, Prague, Czech Republic, September 23-27, 2013, Proceedings, Part III 13 (pp. 387-402). Springer Berlin Heidelberg. 
\bibitem{Lin}
Lin, F. (2021). Supervised learning in neural networks: Feedback-network-free implementation and biological plausibility. IEEE Transactions on Neural Networks and Learning Systems, 33(12), 7888-7898.
\bibitem{Goodfellow}
Goodfellow, I.J., Shlens, J. \& Szegedy, C. (2015) Explaining and harnessing adversarial examples, arXiv:1412.6572v3.
\bibitem{Shafahi}
Shafahi, A., Huang, W. R., Studer, C., Feizi, S., \& Goldstein, T. Are adversarial examples inevitable?,  arXiv:1809.02104v3.
\bibitem{Zhang22}
Zhang, C., Benz, P., Lin, C., Karjauv, A., Wu, J., \& Kweon, I. S. (2021). A survey on universal adversarial attack, arXiv:2103.01498v2.
\bibitem{Ilyas}
Ilyas, A., Santurkar, S., Tsipras, D., Engstrom, L., Tran, B., \& Madry, A. (2019). Adversarial examples are not bugs, they are features. Advances in neural information processing systems, 32.
\bibitem{Zhang23}
Zhang, X. Y., Xie, G. S., Li, X., Mei, T., \& Liu, C. L. (2023). A survey on learning to reject. Proceedings of the IEEE, 111(2), 185-215.
\bibitem{Schmidt}
Schmidt, L., Santurkar, S., Tsipras, D., Talwar, K., \& Madry, A. (2018). Adversarially robust generalization requires more data. Advances in neural information processing systems, 31.
\bibitem{Fawzi}
Fawzi, A., Moosavi-Dezfooli, S. M., \& Frossard, P. (2016). Robustness of classifiers: from adversarial to random noise. Advances in neural information processing systems, 29.
\bibitem{Crecchi}
 Crecchi, F., Melis, M., Sotgiu, A., Bacciu, D., \& Biggio, B. (2022). Fader: Fast adversarial example rejection. Neurocomputing, 470, 257-268.
\bibitem{Tumer}
Tumer, K., \& Ghosh, J. (1996). Error correlation and error reduction in ensemble classifiers. Connection science, 8(3-4), 385-404.
\bibitem{Windeatt11}
Windeatt, T., \& Zor, C. (2011). Minimising added classification error using Walsh coefficients. IEEE transactions on neural networks, 22(8), 1334-1339.
\bibitem{Windeatt13}
Windeatt, T., \& Zor, C. (2013). Ensemble pruning using spectral coefficients. IEEE transactions on neural networks and learning systems, 24(4), 673-678.
\bibitem{Nicolae}
Nicolae, M. I., Sinn, M., Tran, M. N., Buesser, B., Rawat, A., Wistuba, M., ... \& Edwards, B. (2023) Adversarial Robustness Toolbox, arXiv:1807.01069v4.
\bibitem{Krizhevsky}
Krizhevsky A, CIFAR-10 dataset (2023) [Online]. Available: https://www.cs.toronto.edu/~kriz/cifar.html.
\bibitem{Moosavi16}
Moosavi-Dezfooli, S. M., Fawzi, A., \& Frossard, P. (2016). Deepfool: a simple and accurate method to fool deep neural networks. In Proceedings of the IEEE conference on computer vision and pattern recognition (pp. 2574-2582).
\bibitem{Madry}
Madry, A. (2019). Towards deep learning models resistant to adversarial attacks. Towards deep learning models resistant to adversarial attacks, arXiv:1706.06083v4.
\bibitem{Moosavi17}
Moosavi-Dezfooli, S. M., Fawzi, A., Fawzi, O., \& Frossard, P. (2017). Universal adversarial perturbations. In Proceedings of the IEEE conference on computer vision and pattern recognition (pp. 1765-1773).
\bibitem{Kurakin}
Kurakin, A., Goodfellow, I. J., \& Bengio, S. (2018). Adversarial examples in the physical world. In Artificial intelligence safety and security (pp. 99-112). Chapman and Hall/CRC.
\bibitem{Serban}
Serban, A., Poll, E., \& Visser, J. (2020). Adversarial examples on object recognition: A comprehensive survey. ACM Computing Surveys (CSUR), 53(3), 1-38.
\bibitem{Aldahdooh} 
Aldahdooh, A., Hamidouche, W., Fezza, S. A., \& Déforges, O. (2022). Adversarial example detection for DNN models: A review and experimental comparison. Artificial Intelligence Review, 55(6), 4403-4462.
\bibitem{Kwon}
Kwon, H., Kim, Y., Yoon, H., \& Choi, D. (2021). Classification score approach for detecting adversarial example in deep neural network. Multimedia Tools and Applications, 80, 10339-10360.
\bibitem{Pang}
Pang, T., Xu, K., Du, C., Chen, N., \& Zhu, J. (2019, May). Improving adversarial robustness via promoting ensemble diversity. In International Conference on Machine Learning (pp. 4970-4979). PMLR.
\bibitem{Sen}
Sen, S., Ravindran, B. \&  Raghunathan, A. (2020) Ensembles of mixed precision deep networks for increased robustness against adversarial attacks, arXiv:2004.10162v1.
\bibitem{Verma}
Verma, G., \& Swami, A. (2019). Error correcting output codes improve probability estimation and adversarial robustness of deep neural networks. Advances in Neural Information Processing Systems, 32.
\bibitem{LeCun}
LeCun Y, MNIST dataset (2023) [0nline] Available: http://yann.lecun.com/exdb/mnist/.
\bibitem{Krizhevsky12}
Krizhevsky, A., Sutskever, I., \& Hinton, G. E. (2012). Imagenet classification with deep convolutional neural networks. Advances in neural information processing systems, 25.





\end{thebibliography}
\end{document}